\documentclass[conference]{IEEEtran}
\IEEEoverridecommandlockouts
\usepackage{cite}
\usepackage{amsmath,amssymb,amsfonts}
\usepackage{algorithmic}
\usepackage{graphicx}
\usepackage{textcomp}
\usepackage{xcolor}
\usepackage{booktabs}
\usepackage{multirow}
\usepackage{multicol}
\usepackage{makecell}
\def\BibTeX{{\rm B\kern-.05em{\sc i\kern-.025em b}\kern-.08em
    T\kern-.1667em\lower.7ex\hbox{E}\kern-.125emX}}
\begin{document}

\title{Boundary-Aware Proposal Generation Method for \\ Temporal Action Localization
}

\author{Hao Zhang, Chunyan Feng, Jiahui Yang, Zheng Li, Caili Guo\\
	Beijing University of Posts and Telecommunications \\
	{\tt\small 
		\{zhanghao0215, cyfeng, Yangjh, lizhengzachary, guocaili \}@bupt.edu.cn} \\
}
\maketitle
\begin{abstract}
The goal of Temporal Action Localization (TAL) is to find the categories and temporal boundaries of actions in an untrimmed video. Most TAL methods rely heavily on action recognition models that are sensitive to action labels rather than temporal boundaries. More importantly, few works consider the background frames that are similar to action frames in pixels but dissimilar in semantics, which also leads to inaccurate temporal boundaries. To address the challenge above, we propose a Boundary-Aware Proposal Generation (BAPG) method with contrastive learning. Specifically, we define the above background frames as hard negative samples. Contrastive learning with hard negative mining is introduced to improve the discrimination of BAPG. BAPG is independent of the existing TAL network architecture, so it can be applied plug-and-play to mainstream TAL models. Extensive experimental results on THUMOS14 and ActivityNet-1.3 demonstrate that BAPG can significantly improve the performance of TAL.
\end{abstract}


\section{Introduction}
Temporal action localization (TAL) aims to identify the class and duration of actions in an untrimmed video, which is of great importance to video understanding. Thanks to the import of various advanced deep learning networks, such as temporal convolution networks~\cite{dai2022mstct}, pyramid networks~\cite{shi2023tridet,zhao2017ssn}, and Transformers~\cite{zhang2022actionformer}, the performance of TAL has been greatly improved. However, due to the rich spatio-temporal information in videos, TAL still remains a very challenging task. A crucial challenge in TAL is how to accurately locate the temporal boundaries, where the action of interest occurs.

\begin{figure}[t]
	\centerline{\includegraphics[width=\linewidth]{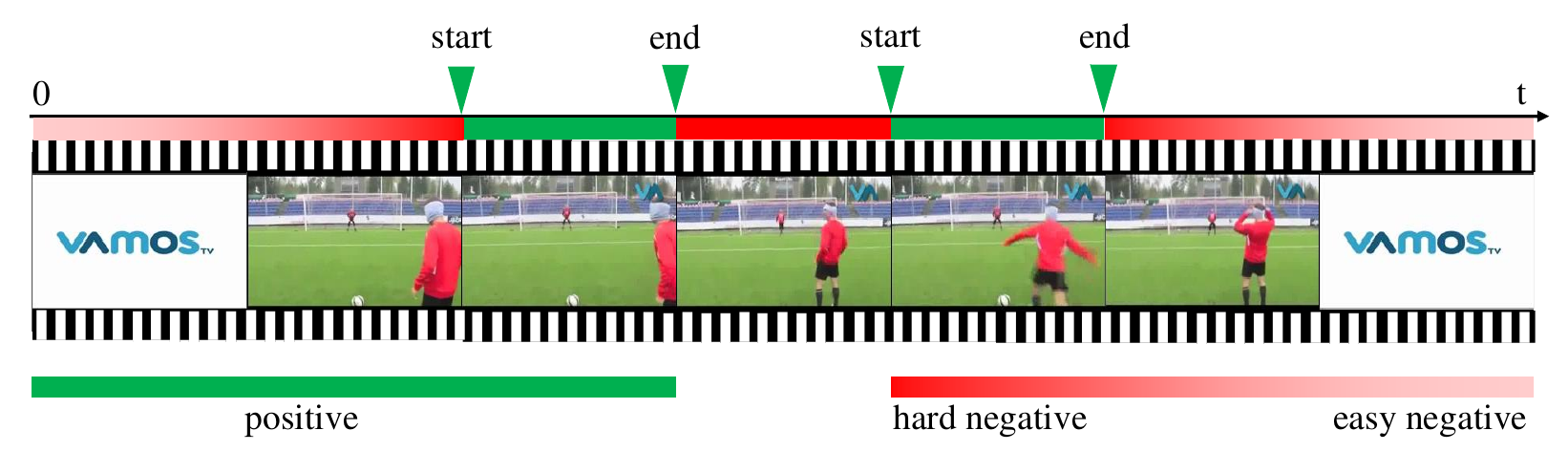}}
	\caption{Illustration of all sample species in the video. The positive samples (green) and easy negative samples (pink) can be easily identified by classification networks. The hard negative samples (red) are usually near the ground-truth temporal boundaries in time. Since the hard negative samples are similar in pixels but dissimilar in semantics to the positive samples, they are very difficult to identify and raise issues with inaccurate temporal boundaries.} 
	\label{fig:motivation}
\end{figure}

Existing methods can be divided into two categories: segment-based methods~\cite{lin2018bsn, Lin2019BMN, Long2019GTAN, Lin2021AFSD, zhang2022actionformer, shi2023tridet, liu2022tadtr} and frame-based methods~\cite{Liu2019MGG, cheng2022tallformer, kang2023AMNet}. Most segment-based methods obtain the temporal boundaries, i.e., start and end timestamps, by using the action recognition network to predict the probability of action occurrence. However, this method is limited in capturing temporal boundaries, because boundary annotations are not available during training~\cite{liu2016ssd, redmon2016yolo, redmon2017yolo9000, lin2017SSAD, tran2015c3d, kim20141dcnn, carreira2017i3d}. In other words, the action classification networks are sensitive to action classes rather than temporal boundaries. The other kinds of methods take video frames into consideration, but most of them~\cite{kang2023AMNet} ignore the hard negative samples. As shown in Fig.~\ref{fig:motivation}, hard negative samples in TAL refer to background frames with similar background information as ground-truth action frames but with different semantics. Due to misclassification of hard negative samples, frame-based methods can lead to inaccurate start and end time timestamps, i.e., inaccurate temporal boundaries.

\begin{figure*}[t]
	\includegraphics[width=\linewidth]{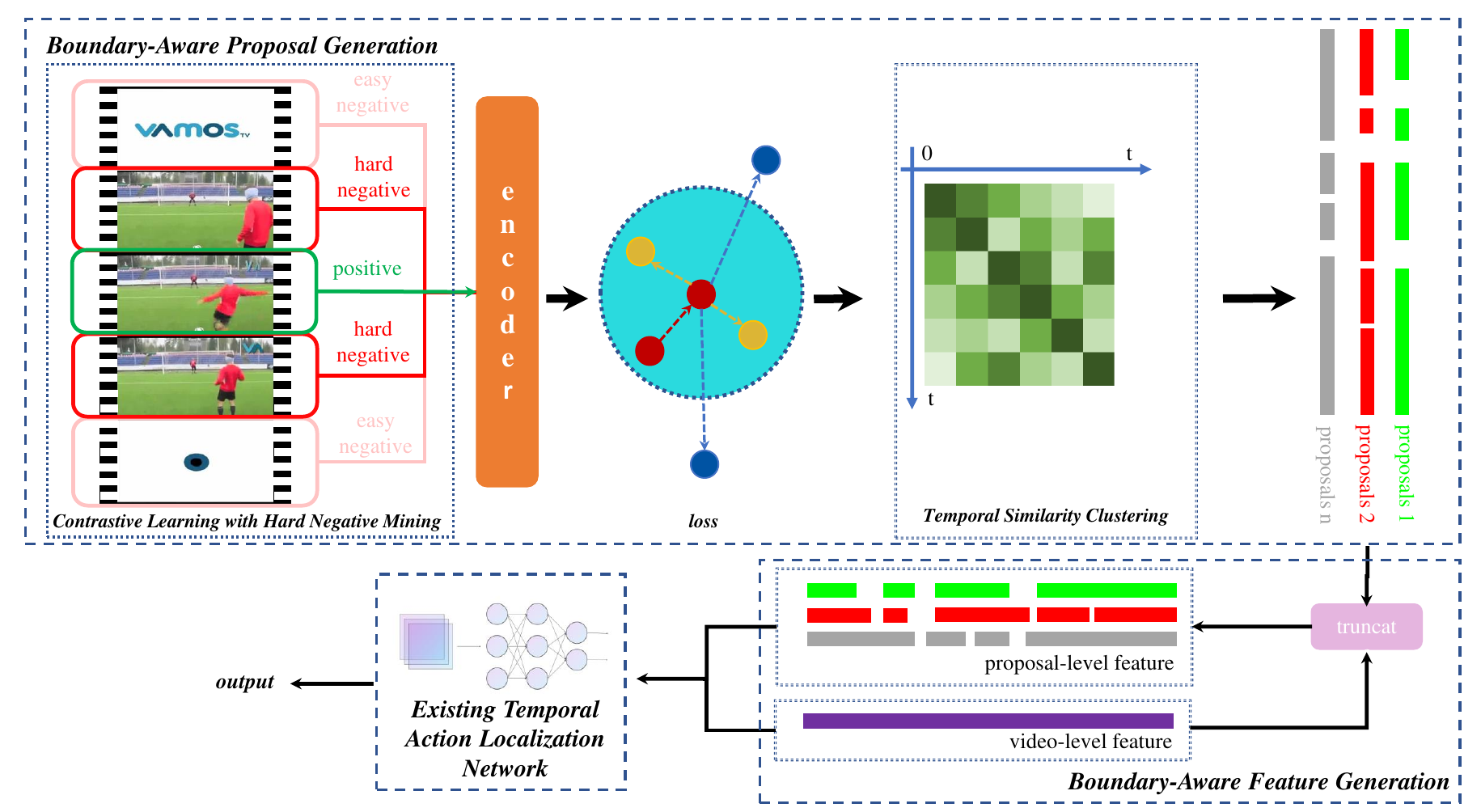}
	\caption{The overall framework of BAPG. In contrastive learning module, we design a novel hard negative mining strategy to make the distance between positive pairs close, and negative pairs far away. The temporal similarity clustering is proposed to generate proposals with high similarities in the proposal generation module. Finally, by combining the proposal-level features with the video-level features, we can promote the performance of existing TAL methods.}
	\label{fig:framework}
\end{figure*}

We propose a novel Boundary-Aware Proposal Generation (BAPG) method with contrastive learning to generate accurate temporal boundaries. As Fig.~\ref{fig:framework} shown, BAPG contains contrastive learning module, proposal generation module and multi-scale feature generation module. In contrastive learning module, we introduce contrastive learning with hard negative mining to measure the similarity of all the frames in each video in~\ref{sec:cl}. Unlike existing methods that rely heavily on action recognition networks, the contrastive learning module can improve the discrimination ability of BAPG by distinguishing hard negative samples. Based on the similarities, a novel proposal generation methods called temporal similarity clustering is proposed to generate fine-grained temporal boundaries in~\ref{sec:tsc}. Using the generated proposal, we can get the boundary-sensitive features in ~\ref{sec:bafg}. In particular, we combine the features of proposal-level and original video-level as the input of the existing action localization model to obtain the classification and localization results. As a plug-and-play method, BAPG is independent of the existing TAL network architecture and can seamlessly connect with existing TAL models. Subsequent experiments verified its effectiveness.

\indent Our contributions can be summarized as follows:
\begin{itemize}
	\item To solve the problem that the existing TAL methods can not obtain accurate temporal boundaries, we propose a novel boundary-aware proposal generation method. The contrastive learning with hard negative mining enables the model to focus on temporal boundaries rather than action categories.
	\item To generate boundary-aware proposals, we propose temporal similarity clustering to capture frames that are similar and temporally continuous.
	
	\item Experiments on two benchmark datasets verify that BAPG can improve the performance of existing TAL networks. 
\end{itemize}
\section{Methodology}

\subsection{Problem Definition}
An input untrimmed video can be denoted as $\mathcal{X}$. The ground-truth annotation is a set of action instances $\Psi = \{\psi_n = (t_{s,n}, t_{e,n}, c_n) \}_{n=1}^{N}$, where $t_{s,n}$, $t_{e,n}$ and $c_n$ represent the start time, the end time and the label of an action instance $\psi_n$ respectively. $N$ is the number of ground-truth action instances in video $X$. TAL aims at detecting all action instances $\Psi$ based on the input video $\mathcal{X}$.
\subsection{Contrastive Learning with Hard Negative Mining}
\label{sec:cl}

\begin{figure}[t]
	\includegraphics[width=\linewidth]{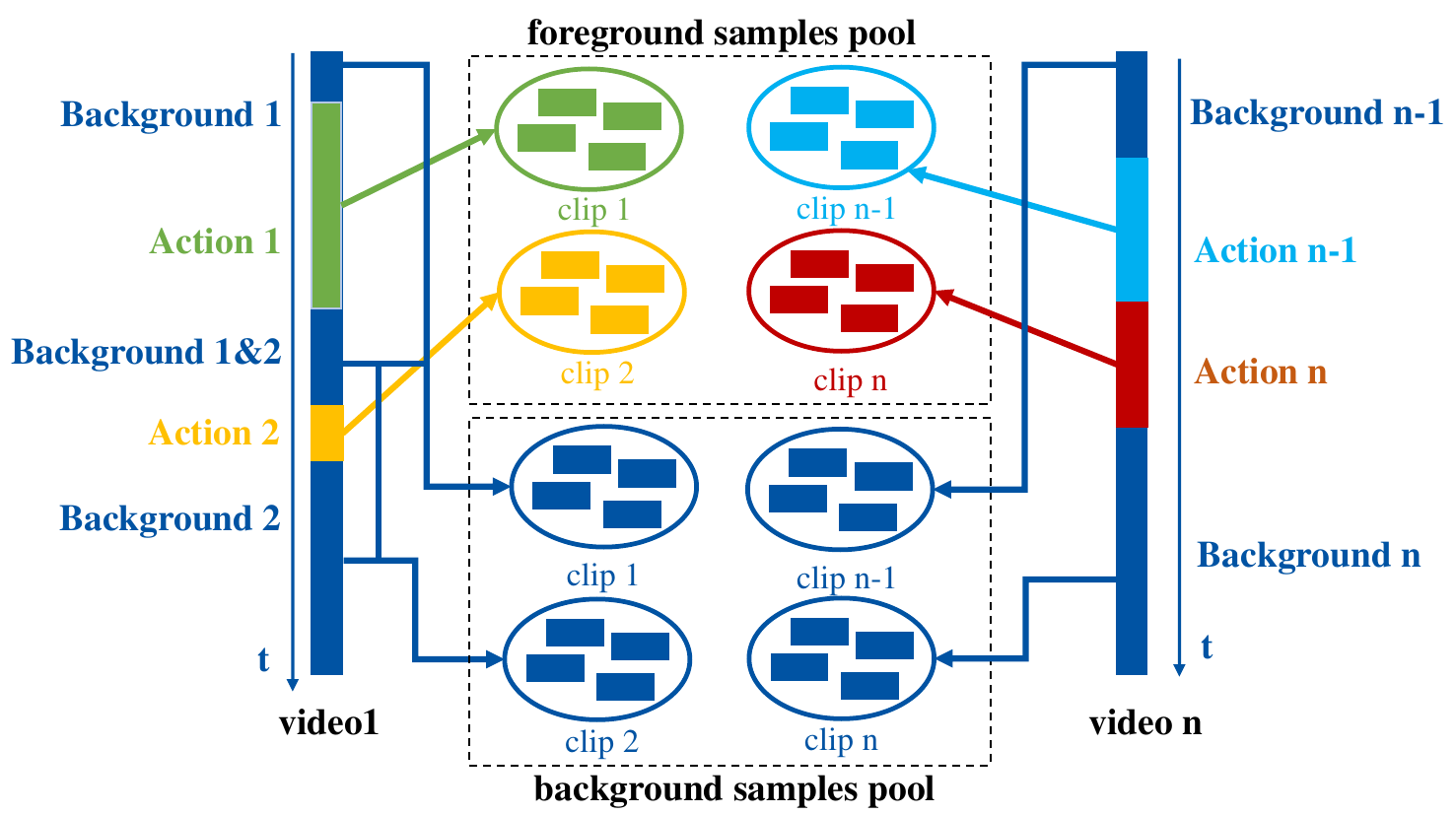}
	\caption{The illustration of positive and negative sample pool construction.}
	\label{fig:sample_pool}
\end{figure}

We first uniformly sample the video $\mathcal{X}$ at time interval $T$ to get $l$ frames $\{X_i\}_{i=1}^{l}$. We take each ground-truth frame as an anchor and denote it as $X_a$. A positive sample $X_p$ is a video frame that belong to the same action instance as the anchor. Otherwise, the sample is negative sample $X_n$. An anchor and a positive sample form a positive pair. Similarly, an anchor and a negative sample construct a negative pair. The hard negative sample $X_{hn}$ refers to a negative sample that is easily confused with a positive sample, which causes boundary confusion. Otherwise, the negative sample is easy negative sample $X_{en}$. To alleviate the temporal boundary confusion problem prevalent in the existing TAL network, we introduce contrastive learning to make the distance between positive pairs close, and negative pairs far away.

\noindent\textbf{Hard Negative Mining.}The positive sample $X_p$ and easy negative sample $X_{en}$ can be easily detected by a binary classifier in existing methods to build up the temporal boundaries. But most of these methods ignore the hard negative sample $X_{hn}$, which have an impact on the start and end timestamps of proposals. Although the hard negative samples are similar in pixels to the positive samples in Fig.~\ref{fig:framework}, the ground-truth clip only includes the positive foreground, and the red area is the warm-up or cool-down of the interested action. Ignoring hard negative samples leads to inaccurate temopral boundaries.

Based on the above reasoning, we design a new hard negative mining strategy that can efficiently distinguish hard negative samples during training. According to the ground-truth, we divide the video into clips that stand for different meaning. As shown in Fig.~\ref{fig:sample_pool}, each video can be split into background clips and action clips. We take the frames in action clips as positive samples, which constitute the positive sample set. The former and latter background clip of the action clip is defined as the hard negative clip. The frames taken from the hard negative clip are the hard negative samples, which constitute the hard negative sample set.

\noindent\textbf{Training and Inference.} As a plug and play method, BAPG only needs to be trained in the contrastive learning phase. For each training, we select an anchor and a positive sample from the positive sample set. Likewise, we select an hard negative sample from the corresponding hard negative sample set. We train a CNN $\Phi(\cdot)$ as the encoder of contrastive learning to obtain the similarity of different kinds of frames. Then, the CNN is employed to extract the feature of each sample $X_i$:
\begin{equation}
	\mathbf{x}_i = \Phi(X_i).
\end{equation}
The similarity between samples is calculated by cosine similarity:
\begin{equation}
	s(X_a,X_p) = \dfrac{\mathbf{x}_a^\top \cdot \mathbf{x}_p}{||\mathbf{x}_a|| \cdot ||\mathbf{x}_p||}
\end{equation}
The encoder is trained to minimize:
\begin{equation}
		\mathcal{L} =  [s(X_a,X_n)-\gamma ]_{+} - s(X_a,X_p)
\end{equation}
where $ [x]_+ = \max(x, 0) $, $\gamma$ denotes the predifined margin (set to 1 as common sense). The margin ensures that embeddings are not projected arbitrarily far apart from each other. The loss function can increase the similarity of positive sample pairs and decrease the similarity of negative sample pairs.

\indent The similarity matrix  $\mathbf{S} = [s(X_i,X_j)]$ of all the frames in a video can be derived during inference. The following $s(X_i, X_j)$ is simplified as $s_{i,j}$.


\subsection{Proposal generation with Temporal Similarity Clustering}
\label{sec:tsc}
Based on the similarity matrix $\mathbf{S}$ obtained by contrastive learning, we need to explore a fast and efficient method for generating precise boundary-aware proposals.

Inspired by~\cite{potapov2014category}, we propose Temporal Similarity Clustering (TSC) method based on change point~\cite{harchaoui2008kernel, harchaoui2007retrospective} to generate fine-grained proposals in units of frames. Compared with the existing proposal generation methods that rely heavily on the accuracy of motion recognition, TSC can generate proposals with precise start and end timestamps, regardless of action categories.\\
\indent\textbf{Temporal Similarity Clustering (TSC).} First, given the similarity matrix $\mathbf{S}$ obtained by contrastive learning, the clip variances are computed for each frame. Following~\cite{crow1984summed}, the variances $v_{t_i, t_{i+1}}$ between $t_{i+1}$ and $t_i$ can be obtained by precomputing the cumulative sums of the similarity matrix:
\begin{equation}
	v_{t_i, t_{i+1}} = \sum_{i=t_i}^{t_{i+1}} s_{i,j} - \sum_{i,j=t_i}^{t_{i+1}} \frac{s_{i,j}}{t_{i+1}-t_i}
\end{equation}
where $s_{i,j}$ is the similarity between $i$-th frame and $j$-th frame in $\mathbf{S}$. 

Then, we employ the dynamic programming algorithm to minimize the best objective value for $i$-th frame and $j$-th frame iteratively:
\begin{equation}
	s_{t_i, t_{i+1}} = {\min}_{t=i,...,j}(s_{i-1,t} + s_{t,j})
\end{equation}
\indent In this way, we can use TSC to divide $\mathbf{S}$ into several regions, the timestamp corresponding to the start frame of each region is $t_s$, and the timestamp corresponding to the end frame is $t_e$. Aiming at generating proposal with different temporal boundaries, we artificially fixed the change point value to a certain set. Finally, through different sets of change points, we can divide the video into diverse clips with different intersection and get various start timestamps $t_s$ and end timestamps $t_e$. Depending on the pair of $t_s$ and $t_e$, we can realize proposal generation.The effects of different change point values can be found in~\ref{sec:ablation} of the ablation experiment.
\subsection{Boundary-Aware Feature Generation}
\label{sec:bafg}
The goal of boundary-aware feature generation is to effectively utilize the original video-level feature $\mathbf{F}_v$ and proposal-level features $\mathbf{F}_p=\{\mathbf{F}_p^1, \mathbf{F}_p^2,...,\mathbf{F}_p^M \}$ to generate precise temporal boundaries, where $M$ is the number of proposals.\\ 
\indent Generally, we extract the video-level feature $\mathbf{F}_v$ through I3D network~\cite{carreira2017i3d}. Now that various proposals have been obtained, we can get the proposal-level features $\mathbf{F}_p$ based on the original video-level feature $\mathbf{F}_v$. Specifically, we use $t_s$ and $t_e$ to get the start timestamp and end timestamp of this proposal. After mapping the exact timestamp to the original video-level feature $\mathbf{F}_v$, we can determine which clip it belongs to and then truncate the video-level feature $\mathbf{F}_v$ corresponding to the time span to obtain the proposal-level features $\mathbf{F}_p$. The proposal-level features $\mathbf{F}_p$ focuses on local description, but pays more attention to the temporal boundaries of an action than the video-level feature $\mathbf{F}_v$. Therefore, we combine the features of two different scales as the input of the existing action localization network, and use the proposal-level features $\mathbf{F}_p$ to refine the video-level feature $\mathbf{F}_v$, which can balance the action temporal boundary and category as well as provide local and global perspectives.

\renewcommand\arraystretch{1.2}

\begin{table}[htbp]
	\caption{Performance comparison on THUMOS14 in terms of mAP at different IoU thresholds. The ``Avg'' column denotes the average mAP in $[0.3:0.1:0.7]$.}
\begin{center}
\begin{tabular}{c|cccccc}
\toprule 
\multirow{2}{*}{Model} & \multicolumn{6}{c}{THUMOS14~(\%)}  \\ 
\cline{2-7} 
& 0.3   & 0.4   & 0.5   & 0.6   & 0.7   & Avg.  \\
			
\midrule
TadTR~\cite{liu2022tadtr}& 74.80 & 69.11 & 60.10 & 46.63 & 32.84 & 56.69 \\
+ \textbf{BAPG} & \textbf{75.44} & \textbf{69.85} & \textbf{60.86} & \textbf{47.34} & \textbf{33.31} & \textbf{57.36}\\

\midrule
Actionformer~\cite{zhang2022actionformer}& 81.96 & 77.41 & 71.08 & 58.91 & 43.74 & 66.62 \\
+ \textbf{BAPG} & \textbf{82.09} & \textbf{77.66} & \textbf{71.47} & \textbf{59.51} & \textbf{44.39} & \textbf{67.02} \\

\midrule 
TriDet~\cite{shi2023tridet}& 83.53 & 79.60 & 72.12 & 60.76 & 45.40 & 68.28\\

+ \textbf{BAPG} & \textbf{83.60} & \textbf{79.72} & \textbf{72.49} & \textbf{61.49} & \textbf{46.35} & \textbf{68.73}\\
\bottomrule 

\end{tabular}
\label{tab:sota_thumos}
\end{center}
\end{table}

\section{Experiments}

\subsection{Experimental Settings}
\textbf{THUMOS14}~\cite{jiang2014thumos} consists of 200 validation videos and 213 test videos for temporal action localization. Without loss of generality, we apply training on the validation subset and evaluate model performance on the test subset~\cite{zeng2021graph}. \textbf{ActivityNet-1.3}~\cite{caba2015activitynet} contains 10,024 videos and 15,410 action instances for training, 4,926 videos and 7,654 action instances for validation and 5044 videos for testing. Following the standard practice~\cite{liu2021multi}, we train BAPG on the training subset and test it on the validation subset.

\subsection{Implementation Details}
In the training phase of contrastive learning, we sample the input data from the foreground and background sample pools according to the designed mask. The encoder $\phi_s$ used for feature extraction is ResNet-50. We train $\phi_s$ for 100 epochs with batch size $b=112$. Following~\cite{RevisitDML,RevisitDML1}, we set the learning rate to $0.00001$ and decays in the multi-step schedule. The optimizer is Adam and the weight decay is $0.0004$. The number of samples selected from one class $n=2$. More parameter settings can be found in the Ablation Experiment section~\ref{sec:ablation}.

\subsection{Improving State-of-the-Art Methods}
\noindent\textbf{THUMOS14.} The experiment results of our BAPG with the state-of-the-art method in THUMOS14 are shown in Table~\ref{tab:sota_thumos}. The table shows that the performance improvement will be increasingly noticeable as tIoU increases. Especially when IoU = 0.7, the mAP can be improved by $2\%$. The experiment results show that BAPG is boundary-sensitive and effective.
\renewcommand\arraystretch{1.2}

\begin{table}[t]
\caption{Performance comparison on ActivityNet v1.3 in terms of mAP at different IoU thresholds. The ``Avg'' columns denote the average mAP in $[0.5:0.05:0.95]$.}
\begin{center}
\begin{tabular}{c|ccccc}
\toprule
\multirow{2}{*}{Model} & \multicolumn{5}{c}{ActivityNet v1.3~(\%)}  \\ 
\cline{2-6} 
& 0.5  & 0.6  & 0.7 & 0.8 & Avg.  \\
			

%
\midrule
Actionformer~\cite{zhang2022actionformer}& 54.67 & 48.25 & 41.85 & 32.91 & 36.56 \\
+ \textbf{BAPG} & \textbf{54.80} & \textbf{48.38} & \textbf{41.97} & \textbf{32.93} & \textbf{36.61}\\

\midrule 
TriDet~\cite{shi2023tridet}& 54.71 & 48.54 & 42.34 & 32.93 & 36.76\\
+ \textbf{BAPG} & \textbf{54.85} & \textbf{48.58} & \textbf{42.50} & \textbf{33.15} & \textbf{36.83} \\
\bottomrule
\end{tabular}
\label{tab:sota_anet}
\end{center}
\end{table}

\noindent\textbf{ActivityNet-1.3.} The experiment results of our BAPG with the state-of-the-art method in ActivityNet-1.3 is shown in Table~\ref{tab:sota_anet}. Although the videos in ActivityNet-1.3 are more complicated and variable than THUMOS14, BAPG can still improve the performance of existing methods by $0.2\%$.
\subsection{Ablation Study on Hyper-Parameters}
\label{sec:ablation}
In this section, we conduct ablation experiments on the THUMOS14 dataset to verify the effect of hyper-parameters on BAPG performance. The baseline method is actionformer~\cite{zhang2022actionformer}.

\renewcommand\arraystretch{1.2}

\begin{table}[htbp]
\caption{Effect of different video sampling rates.}
\begin{center}
\begin{tabular}{c|c|ccccccc}
\toprule

\multirow{2}{*}{Method} &\multirow{2}{*}{Rate(s/f)}& \multicolumn{6}{c}{mAP@tIoU (\%)} \\
\cline{3-8} 
& & 0.3  & 0.4 & 0.5 & 0.6 & 0.7 & Avg. \\ 

\midrule 
baseline & - & 81.96 & 77.41 & 71.08 & 58.91 & 43.74 & 66.62 \\

\midrule 
\multirow {1}{*}{BAPG} &{0.1} & \textbf{82.09} & \textbf{77.66} & \textbf{71.47} & \textbf{59.51} & \textbf{44.39} & \textbf{67.02}    \\

\midrule 
\multirow {1}{*}{BAPG} &{1} & 82.05 & 77.62 & 71.44 & 59.37 & 44.36 & 66.97 \\

\bottomrule 
		\end{tabular}
		\label{tab:sampling rate}
	\end{center}
\end{table}

\noindent\textbf{Different video sampling rate.} To measure the impact of different time spans on contrastive learning, we used two video sampling methods, namely $1s$ per frame and $0.1s$ per frame. It can be clearly seen from table~\ref{tab:sampling rate} that the performance of BAPG with a video sampling rate of $0.1s$ per frame is fully ahead of $1s$ per frame and the baseline method. The results show that the bigger the sampling rate, the more comprehensive the similarity that the network can learn and the more robust to proposals of different overlaps. Considering the trade-off between implementation complexity and precision, we choose the sampling rate of $1s$ per frame.
\renewcommand\arraystretch{1.2}

\begin{table}[htbp]
\caption{Effect of different change points.}
\begin{center}
\begin{tabular}{c|cccccc}
\toprule
\multirow{2}{*}{Change Point} & \multicolumn{6}{c}{mAP@tIoU~(\%)}  \\ 
\cline{2-7} 
& 0.3  & 0.4 & 0.5 & 0.6 & 0.7 & Avg. \\
\midrule 
Baseline& 81.96 & 77.41 & 71.08 & 58.91 & 43.74 & 66.62 \\
			
\midrule 
\multirow {1}{*}{BAPG($m=10$)} & 82.05 & 77.62 & 71.44 & 59.37 & 44.36 & 66.97 \\

\midrule 
\multirow {1}{*}{BAPG($m=15$)} & \textbf{82.08} & \textbf{77.64} & \textbf{71.45} & 59.39 & 44.41 & 67.00 \\

\midrule 
\multirow {1}{*}{BAPG($m=20$)} & 82.06 & 77.63 & \textbf{71.45} & \textbf{59.44} & \textbf{44.49} & \textbf{67.01} \\		

\bottomrule
\end{tabular}
\label{tab:changepoint}
\end{center}
\end{table}

\noindent\textbf{Different change point value.} Different change point values affect the number of proposals generated in each video. The more the change point value, the more proposal scales are generated. It can be seen from the table ~\ref{tab:changepoint} that the sparse change point value can achieve better performance when the tIoU is small, and the dense change point value can significantly improve the performance when the tIoU is large. Considering the trade-off between implementation complexity and precision, we choose $m = 10$.

\section{Conclusion}
In this paper, we propose a novel plug-and-play method called BAPG to generate boundary-aware proposals. BAPG aims to introduce contrastive learning with hard negative mining to solve a major challenge: existing networks rely heavily on the performance of action recognition networks and ignore hard negative samples. Experiments on THUMOS14 and ActivityNet-1.3 verify the effectiveness of the proposed method.

\clearpage

\end{document}